\pdfoutput=1

\documentclass[11pt]{article}
\usepackage{booktabs}
\usepackage[final]{acl}
\usepackage[most]{tcolorbox} 
\tcbuselibrary{listingsutf8} 
\usepackage{changepage}  
\usepackage{pgfplots}
\pgfplotsset{compat=1.18}
\usepackage{subcaption}
\usepackage{amsmath}
\usepackage{times}
\usepackage{placeins}
\usepackage{latexsym}
\usepackage{amsmath} 
\usepackage[T1]{fontenc}
\usepackage[utf8]{inputenc}
\usepackage{subcaption}
\usepackage{microtype}
\usepackage{inconsolata}
\usepackage{graphicx}
\usepackage{pgfplots}
\pgfplotsset{compat=1.18}
\usepackage{subcaption}  

\title{When Inverse Data Outperforms: Exploring the Pitfalls of \\ Mixed Data in Multi-Stage Fine-Tuning}
\author{
  Mengyi Deng\thanks{Equal contribution.}$^{1}$,
  Xin Li\footnotemark[1]$^{1}$,
  Tingyu Zhu$^{1}$,
  Zhicheng Yang$^{1}$,
  Zhijiang Guo\thanks{Co-corresponding authors.}$^{1, 2}$,
  Wei Wang\footnotemark[2]$^{1,2}$ \\
  $^{1}$Information Hub, Data Science and Analysis Thrust, HKUST (Guangzhou), China \\
  $^{2}$The Hong Kong University of Science and Technology, Hong Kong SAR, China \\
  \texttt{\{mdeng974, tzhu619, zyang398\}@connect.hkust-gz.edu.cn, lixin2002cn@gmail.com,} \\
  \texttt{zhijiangguo@hkust-gz.edu.cn, weiwcs@hkust-gz.edu.cn, weiwcs@ust.hk}
}

\begin{document}
\maketitle
\begin{abstract}

Existing work has shown that o1-level performance can be achieved with limited data distillation, but most existing methods focus on unidirectional supervised fine-tuning (SFT), overlooking the intricate interplay between diverse reasoning patterns. In this paper, we construct \textbf{r1k}, a high-quality reverse reasoning dataset derived by inverting 1,000 forward examples from s1k~\cite{muennighoff2025s1}, and examine how SFT and Direct Preference Optimization (DPO) affect alignment under bidirectional reasoning objectives. SFT on r1k yields a \textbf{1.6\%--6.8\%} accuracy improvement over s1k across evaluated benchmarks. However, naively mixing forward and reverse data during SFT weakens the directional distinction. Although DPO can partially recover this distinction, it also suppresses less preferred reasoning paths by shifting the probability mass toward irrelevant outputs. These findings suggest that mixed reasoning data introduce conflicting supervision signals, underscoring the need for robust and direction-aware alignment strategies. Our code and data are available at: \url{https://github.com/16demi/ReasonAlign-analysis}.

\end{abstract}

\section{Introduction}

\begin{figure*}
    \centering
    \includegraphics[width=1\linewidth]{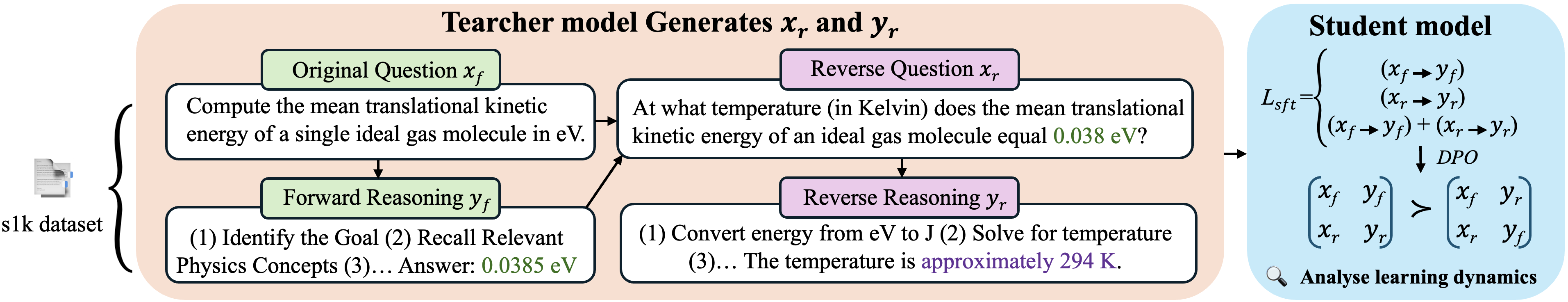}
    \caption{We begin with the s1k dataset $(x_f, y_f)$ and generate reverse questions $x_r$, along with their corresponding reverse CoTs and answers $y_r$. We then fine-tune student models using cross-entropy loss under three settings as comparison: forward-only data, reverse-only data, and a mixture of both. To enhance directional consistency, we apply DPO to encourage directionally aligned responses while suppressing misaligned ones. Concurrently, we track the log probability of $y_f$ and $y_r$ across multiple fine-tuning stages to investigate the models' learning dynamics.
}
    \label{1}
\end{figure*}

Recent studies show that Large Language Models (LLMs) can achieve strong reasoning performance by distilling knowledge from a small set of high-quality examples~\citep{ReasonSurvey25}. Methods like s1~\cite{muennighoff2025s1} and LIMO~\cite{ye2025limo} demonstrate that with just 817 to 1,000 curated samples, a 32B model can match or surpass larger systems. However, these approaches focus mainly on single-direction reasoning—solving problems step by step from question to answer. As shown in Figure~\ref{1}, a model may learn to compute the kinetic energy of a gas molecule from its temperature, but not the reverse: inferring temperature from energy.

This narrow focus overlooks a core aspect of human cognition: \textbf{the inherently bidirectional nature} of reasoning. Humans commonly engage in backward reasoning, particularly in goal-directed problem solving~\cite{newell1972human,hawes2012experience}. Rather than reasoning solely from premises to conclusions, people often begin with a desired outcome and work backward through intermediate steps to reach known facts~\cite{senn2015backward}. Motivated by this cognitive insight, recent studies have begun to explore reverse or backward reasoning in LLM. MathGenie~\cite{li2024mathgenie} utilizes reverse derivation paths to improve robustness on math word problems. Iterative Question Composing~\citep{Iter25} constructs intermediate subquestions that align with goal-driven, backward-style planning. In optimization modeling, OptiBench~\cite{yang2024optibench} promotes reflective and Socratic-style reformulations that partially embody reverse reasoning principles. While promising, these approaches remain constrained to short-context reasoning or domain-specific tasks. This leaves open the broader question of whether backward supervision can enhance long chain-of-thought~\citep{CoT22} reasoning and generalize across diverse scenarios.

To investigate this, we construct a high-quality dataset, \textbf{r1k}, by systematically inverting 1,000 forward reasoning examples from s1k~\cite{muennighoff2025s1}. Reverse questions and reasoning paths are generated with cost-efficient DeepSeek-R1 model~\cite{guo2025deepseek}, without the need for expensive data collection, cleaning, or selection procedures. Fine-tuning on \textbf{r1k} yields an approximate 1.6\%--6.8\% improvement over s1k.

To further study the interaction between mixed data, we conducted extensive experiments on their combined effects. We observe that SFT on reverse data improves performance, whereas mixing forward and reverse examples leads to degradation. Mechanistic analysis shows that this reduces the model’s ability to distinguish reasoning paths. While Direct Preference Optimization (DPO;~\citealt{DPO2023}) partly alleviates this, it still suffers from suboptimal initialization and tends to shift reverse reasoning probability toward irrelevant outputs. These findings underscore the need for improved alignment strategies to support robust reasoning.

\section{Related work}
\textbf{Data-Efficient Reasoning in LLMs:}
s1~\cite{muennighoff2025s1}, LIMO~\cite{ye2025limo} and LIMA~\cite{Zhou2023LIMA} demonstrate that training on a small set of high-quality examples enables more effective performance, suggesting that massive datasets may not always be necessary to achieve competitive results. This perspective is further supported by methods such as iterative refinement~\cite{madaan2023selfrefine} and self-rewarding feedback~\cite{huang2023selfrag}, which demonstrate that reusing or distilling informative examples can improve model performance without relying on large-scale data. Complementary findings from data pruning and selection studies~\cite{agarwal2024delift,huang2024swiftcoder,redstar25} reveal that indiscriminate scaling often yields diminishing returns, highlighting the value of targeted data curation in reasoning-intensive tasks.

\noindent\textbf{Learning Dynamics of LLM Fine-Tuning:}Neural Tangent Kernel (NTK) theory~\cite{jacot2018ntk,arora2019harnessing} provides a framework for analyzing the influence of individual training examples during LLM fine-tuning. A gradient-based decomposition was later proposed~\cite{ren2024learning}, approximating the change in model confidence for an output \(y\) on input \(x_o\) after training on a single example \((x_u, y_u)\) as:
\[
\scalebox{0.9}{$
\Delta \log \pi_t(y \mid x_o) \approx -\eta\, A_t(x_o)\, K_t(x_o, x_u)\, G_t(x_u, y_u)
$}
\]
where $K_t$ is the empirical NTK, $G_t$ the gradient, and $A_t$ a scaling factor tied to model certainty. This perspective helps explain interference, hallucination, and memorization~\cite{pruthi2020tracin}. It also explains the diversity collapse~\cite{dang2025diversity}, where correctness optimization concentrates the probability mass on a single reasoning path, limiting diversity. These insights inspire a learning-dynamics perspective on how mixed reasoning data shapes model behavior in multi-stage fine-tuning.

\section{Methodology}
\subsection{Reverse Data Construction and Alignment}
We begin with a forward reasoning dataset $\mathcal{D}_{s1k} = \{(x_f^{(i)}, y_f^{(i)})\}_{i=1}^{1000}$, consisting of 1,000 high-quality examples from the s1k dataset. Each example includes a question $x_f$ and its corresponding CoT and answer $y_f$ generated by Deepseek-R1 (R1). Based on each s1k’s question and final answer, we leverage R1  to construct the reverse dataset $\mathcal{D}_{r1k} = \{(x_r^{(i)}, y_r^{(i)})\}_{i=1}^{1000}$. For each forward example $(x_f, y_f)$, we prompt the model to generate a reverse question $x_r$ that naturally elicits the original reasoning in reverse. Conditioned on $x_r$, we prompt R1 to generate the corresponding reverse reasoning chain and answer $y_r$. A concrete forward–reverse example is provided in Appendix~\ref{sec:example}. We merge the above two datasets to obtain $\mathcal{D} = \mathcal{D}_{s1k} \cup \mathcal{D}_{r1k}$,  which serves to investigate how bidirectional supervision influences model behavior and alignment.

We fine-tune the Qwen2.5-Instruct~\cite{qwen2024} 7B and 14B models using the standard cross-entropy objective, where the input is the question $x$ and the target output  $y$  is the concatenation of the CoT and the final answer with special separation tokens.

Although SFT introduces forward and reverse reasoning, it does not equip LLMs with the ability to switch between two directions. To better align model responses with question directionality, we apply DPO following SFT. For this, we construct preference pairs of the form \((x, y^+, y^-)\), where \(x\) is the question, \(y^+\) is the preferred response, and \(y^-\) is the response of reverse question.  Specifically, for each example \((x_f, y_f) \in \mathcal{D}_{s1k}\), we treat the forward output as the preferred response, i.e., \(y^+ = y_f\), and the corresponding reverse output as the rejected response, \(y^- = y_r\). In contrast, for each example \((x_r, y_r) \in \mathcal{D}_{r1k}\), we assign \(y^+ = y_r\) as the preferred response and \(y^- = y_f\) as the rejected one. Each pair of preferences \((x, y^+, y^-)\) is used to optimize the DPO objective, which encourages the model to prefer \(y^+\) over \(y^-\).

\subsection{Analysis of the Pitfalls of Mixed Data}

To investigate the fine-tuning behavior during both the SFT and DPO stages, we construct a small \emph{probe training set} consisting of 100 examples: 50 forward instances \(\mathcal{D}_f\) and their corresponding 50 reverse counterparts \(\mathcal{D}_r\). The union of the two forms a mixed test dataset \(\mathcal{D}_m = \mathcal{D}_f \cup \mathcal{D}_r\).

Throughout both the SFT and DPO stages, we monitor model behavior by recording intermediate checkpoints and evaluating the average log-probability (ALP) per token for both \(y^+\) and \(y^-\):
\[
\scalebox{0.9}{$
\left\{
\begin{aligned}
\mathrm{ALP}(y^+) &= \frac{1}{N} \sum_{i=1}^N \frac{1}{|y_i^+|} \sum_{t=1}^{|y_i^+|} \log p(y^+_{i,t} \mid x_i, y^+_{i,<t}), \\
\mathrm{ALP}(y^-) &= \frac{1}{N} \sum_{i=1}^N \frac{1}{|y_i^-|} \sum_{t=1}^{|y_i^-|} \log p(y^-_{i,t} \mid x_i, y^-_{i,<t}).
\end{aligned}
\right.
$}
\]
here, \(N\) denotes the number of examples in the \emph{probe testing set}, and \(|y_i^\pm|\) the length of each evaluated output, capped at 1000 tokens. This evaluation window is typically sufficient to capture the divergence between \(y^+\) and \(y^-\). Since the responses are long-form sequences, we normalize log-probability by sequence length to ensure fair comparison. Motivated by recent theoretical analyses of learning dynamics in LLMs~\cite{ren2024learning}, we track the margin:
\[
\scalebox{0.96}{$
\Delta = \mathrm{ALP}(y^+) - \mathrm{ALP}(y^-),$}\]
which serves as an empirical proxy for \(\Delta \log \pi_t(y \mid x_o)\) in the NTK formulation, with ALP reflecting model certainty \(A_t(x_o)\), forward–reverse pairs indicating input similarity \(K_t(x_o, x_u)\), and training supervision contributing gradient signals \(G_t(x_u, y_u)\). 

\begin{figure*}
    \centering
    \includegraphics[width=1\linewidth]{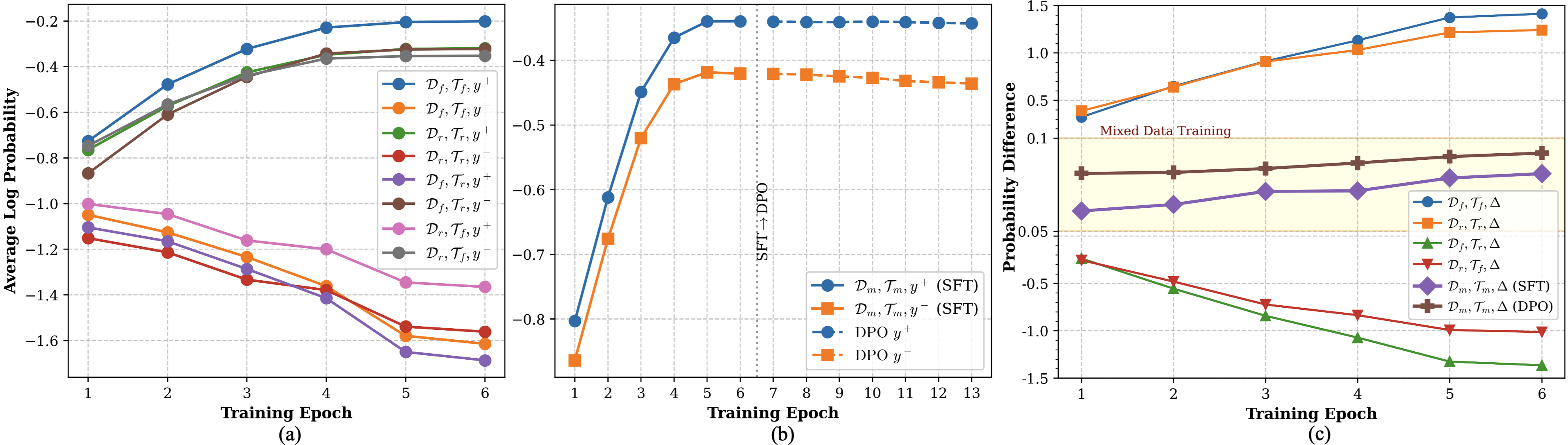}
    \caption{$\mathcal{D}$ denotes the training dataset, and $\mathcal{T}$ denotes the testing dataset. We report the Average Log Probability (ALP) for both the preferred responses ($y^+$) and the less preferred responses ($y^-$) in Figures (a) and (b), respectively. Figure (c) shows the difference $\mathrm{ALP}(y^+) - \mathrm{ALP}(y^-)$.}
    \label{fig:enter-label}
\end{figure*}

\section{Experiments}

We conducted experiments on the DeepSeek-R1's s1k dataset, which consistently outperformed the Gemini-based variant. As the s1k has already been curated with quantity, diversity, and difficulty, we did not apply an additional filtering process. We fine-tune Qwen2.5-Instruct models (7B and 14B) in two stages on 8 A800-80GB GPUs. First, we applied SFT with LoRA (rank = 256, $\alpha$ = 512). Then, we performed DPO using the \texttt{trl} library~\cite{vonwerra2022trl}, with DeepSpeed ZeRO-3~\cite{jacobs2023deepspeed} and Flash Attention~\cite{shah2024flashattention} to reduce memory usage. We employ open-source lm-eval-harness~\cite{eval-harness}, with gpt-4o-mini to evaluate the accuracy.

\subsection{Impact of Reverse and Mixed Data}
To evaluate the effect of reverse data construction and bidirectional supervision, we conducted fine-tuning experiments on different training datasets. As shown in Table~\ref{tab:performance_singlecol}, we compare the distillation performance of models trained on the $\mathcal{D}_{s1k}$, $\mathcal{D}_{r1k}$, and $\mathcal{D}$ across three challenging benchmarks: AIME24-NoFigures~\cite{aime}, Math 500~\cite{lightman2023let}, and GPQA~\cite{rein2024gpqa} benchmarks.
\begin{table}[h]
\centering
\small
\setlength{\abovecaptionskip}{4pt}
\setlength{\belowcaptionskip}{0pt}
\caption{Effect of Reverse Data ($\mathcal{D}_{r1k}$) and Mixed Training Sets ($\mathcal{D}$ and $\mathcal{D}_{0.5k}$) on downstream performance. Here, $\mathcal{D}_{0.5k}$ consists of 500 forward examples paired with their corresponding reverse data. In the same setting experiment, the best results are shown in bold.}
\setlength{\tabcolsep}{3.2pt}  
\begin{tabular}{cccccc}
\toprule
\textbf{Data} & \textbf{Model} & \textbf{AIME} & \textbf{Math} & \textbf{GPQA} & \textbf{Average} \\
\midrule
$\mathcal{D}_{s1k}$ & 3B & 3.3\% & \textbf{45.2\%} & 29.3\% & 25.9\% \\
$\mathcal{D}_{r1k}$(Ours) & 3B & \textbf{6.7\%} & 43.4\% & \textbf{32.2\%} & \textbf{27.5\%} \\
$\mathcal{D}_{0.5k}$ & 3B & 0.0\% & 43.6\% & 29.8\% & 24.47\% \\
$\mathcal{D}$       & 3B & 0.0\% & 43.0\% & 26.2\% & 23.1\% \\
\cmidrule(lr){1-6}
$\mathcal{D}_{s1k}$ & 7B & 16.7\% & 77.0\% & 34.0\% & 42.6\% \\
$\mathcal{D}_{r1k}$(Ours)  & 7B & \textbf{20.0\%} & \textbf{77.4\%} & \textbf{42.4\%} & \textbf{46.6\%}\\
$\mathcal{D}_{0.5k}$ & 7B & 13.3\% & 71.8\% & 35.8\% & 40.3\% \\
$\mathcal{D}$       & 7B & 6.7\%  & 56.0\% & 31.8\%  & 31.5\%\\
\cmidrule(lr){1-6}
$\mathcal{D}_{s1k}$ & 14B & 20.0\% & 83.2\% & 48.4\% & 50.6\% \\
$\mathcal{D}_{r1k}$(Ours) & 14B & \textbf{33.3\%} & \textbf{86.0\%} & \textbf{53.0\%} & \textbf{57.4\%}\\
$\mathcal{D}$       & 14B & 30.0\% & 81.6\% & 49.1\% & 53.6\% \\
\bottomrule
\end{tabular}
\label{tab:performance_singlecol}
\end{table}
\begin{figure*}[t!]
    \centering
    \includegraphics[width=\linewidth]{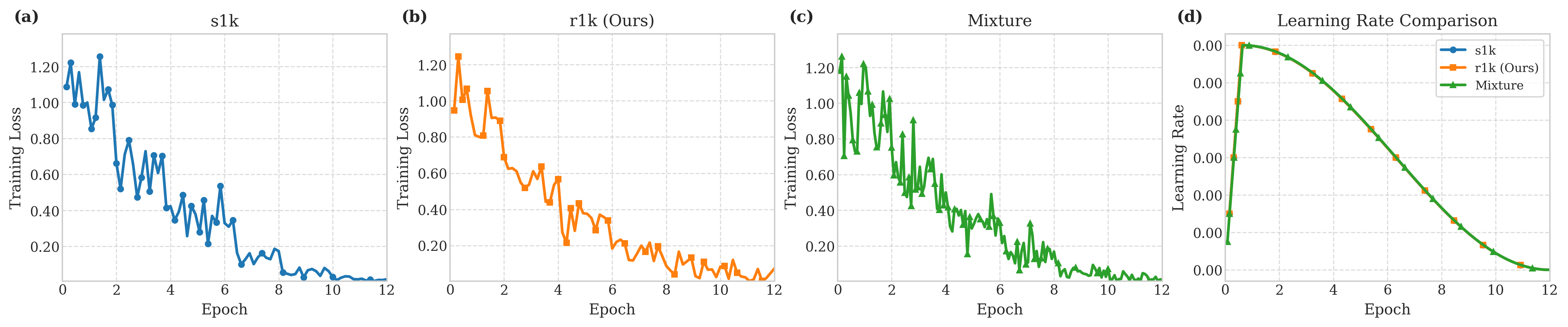}
    \caption{SFT Probe dataset Training Dynamics}
    \label{4}
\end{figure*}

Under the same distillation pipeline, models trained on our reverse dataset $\mathcal{D}_{r1k}$ achieve an average improvement of 1.6\%--6.8\% compared to those trained on the original $\mathcal{D}_{s1k}$. However, combining forward and reverse examples leads to a significant drop in performance. As the size of the mixed dataset increases from \(\mathcal{D}_{0.5k}\) to \(\mathcal{D}\), the degradation becomes more pronounced, suggesting that mixed-direction reasoning data introduce interference between reasoning modes and hinder effective learning.

\subsection{Teacher Model Reliability Analysis}
We further compare SFT results when using Gemini 2.0 Flash Thinking (20–30B) 
versus DeepSeek-R1 (671B MoE) as the teacher model for generating chain-of-thought and answers. 
Table~\ref{tab:appendix_teacher} shows that Gemini-based supervision leads to consistently 1–3\% 
lower downstream accuracy compared to DeepSeek-R1 supervision. All experimental data are sourced from the $\mathcal{D}_{s1k}$ dataset.

\begin{table}[h]
\centering
\small
\setlength{\tabcolsep}{3.2pt}
\caption{SFT based on two teacher models (Gemini and DeepSeek-R1). Parentheses indicate the teacher model used to generate the SFT reasoning and solution data.}
\begin{tabular}{lcccc}
\toprule
\textbf{Model / Teacher} & \textbf{AIME} & \textbf{GPQA} & \textbf{Math} & \textbf{Average} \\
\midrule
7B (Gemini)        & 13.3\% & 33.5\% & 77.4\% & 41.43\% \\
7B (DeepSeek-R1)   & 16.7\% & 34.0\% & 77.0\% & 42.60\% \\
14B (Gemini)       & 16.7\% & 47.2\% & 83.0\% & 48.93\% \\
14B (DeepSeek-R1)  & 20.0\% & 48.4\% & 83.2\% & 50.60\% \\
\bottomrule
\end{tabular}
\label{tab:appendix_teacher}
\end{table}

\subsection{Impact of Directional Preference}

Our DPO experiments use a temperature-weighting hyperparameter of $\beta = 0.6$, with the SFT-trained model fixed as the reference model. We apply DPO fine-tuning to four SFT-based models: three 7B models individually trained on $\mathcal{D}_{s1k}$, $\mathcal{D}_{r1k}$, and $\mathcal{D}$, and a 14B model trained on $\mathcal{D}$. All DPO models are further fine-tuned using preference pairs from $\mathcal{D}$, where each pair consists of two responses $y^+$ and $y^-$, generated from the opposite question.

\begin{table}[h]
\centering
\small
\setlength{\abovecaptionskip}{4pt}
\setlength{\belowcaptionskip}{0pt}
\caption{Effect of DPO on different SFT Data and based models settings. $\downarrow$ and $\uparrow$ indicate performance decrease and increase respectively; parentheses show relative change from the SFT baseline.}
\setlength{\tabcolsep}{3.2pt}
\begin{tabular}{ccccc}
\toprule
\textbf{SFT Data} & \textbf{AIME} & \textbf{Math} & \textbf{GPQA} & \textbf{Average} \\
\midrule
$\mathcal{D}_{s1k}$ (7B) & 13.3\%$\downarrow$ & 71.8\%$\downarrow$ & 35.9\%$\downarrow$ & $40.3\%$ ($\downarrow$2.3\%) \\
$\mathcal{D}_{r1k}$(7B)   & 16.7\%$\downarrow$ & 75.4\%$\downarrow$ & 39.4\%$\downarrow$ & $43.8\%$ ($\downarrow$2.8\%) \\
$\mathcal{D}$ (7B)   & 16.7\%$\uparrow$ & 64.2\%$\uparrow$ & 34.8\%$\uparrow$ & $38.6\%$ ($\uparrow$ 7.1\%)\\
$\mathcal{D}$ (14B)  & 40.0\%$\uparrow$ & 81.2\%$\downarrow$ & 46.4\%$\downarrow$ & $55.9\%$ ($\uparrow$ 2.3\%)\\
\bottomrule
\end{tabular}
\label{tab:train_config_perf}
\end{table}

Table \ref{tab:train_config_perf} shows that applying DPO with mixed preference data on the $\mathcal{D}_{s1k}$ (7B) and $\mathcal{D}_{r1k}$ (7B) reference models leads to a performance decline. For the model initially fine-tuned on the mixed dataset $\mathcal{D}$, DPO achieves some performance improvements, but its overall performance remains inferior to SFT trained on $\mathcal{D}_{r1k}$. 

\subsection{Analysis of the Pitfalls of Mixed Data}

We analyze the in-distribution pairs \((\mathcal{D}_f, \mathcal{T}_f)\), \((\mathcal{D}_r, \mathcal{T}_r)\), and \((\mathcal{D}_m, \mathcal{T}_m)\), where \(\mathcal{D}\) and \(\mathcal{T}\) denote the training and testing datasets respectively. SFT is run for 12 epochs with evaluation every 2 epochs, and DPO for 7 epochs with evaluation after each. Changes in the ALP of $y^+$ reflect the learned strategy, while  $y^-$ indicates hallucination. We also consider the out-of-distribution pairs \((\mathcal{D}_f, \mathcal{T}_r)\) and \((\mathcal{D}_r, \mathcal{T}_f)\), where variations in the ALP of $y^+$ measure the generalization capability to handle reverse question, whereas $y^-$ indicate the likelihood of generating irrelevant or off-target responses.

Figure~\ref{fig:enter-label}(a) shows that under out-of-distribution scenarios, models trained on the \(\mathcal{D}_r\) exhibit lower hallucination rates $y^-$ and better generalization. For in distribution settings, (a) demonstrates that the likelihood of $y^+$ increase significantly, but this improvement is accompanied by a corresponding rise in hallucinations $y^-$. Figure~\ref{fig:enter-label}(b) reveals that mixed-data training \(\mathcal{D}_m\) induces a stronger increase in hallucinations, while the likelihood of preferred responses fails to reach the levels achieved by training solely on $\mathcal{D}_r$ and $\mathcal{D}_f$. Even though the subsequent DPO improves preference alignment by suppressing the probability of $y^-$ to irrelevant responses, this suppression is limited.

Figure~\ref{fig:enter-label}(c) shows that models trained on \(\mathcal{D}_f\) and \(\mathcal{D}_r\) maintain a gap between $y^+$ and $y^-$, whereas the model trained on mixed data \(\mathcal{D}_m\) produces only a narrow margin (0.05–0.1). This suggests that SFT on mixed data weakens LLM’s ability to discriminate the learned strategies and hallucination. Although DPO slightly separates $y^+$ and $y^-$, the effect remains limited. We hypothesize that the conflicting signals from \(\mathcal{D}_m\) lead the model to optimize in competing directions, hindering the formation of coherent preferences. This phenomenon also helps explain why models trained on smaller but higher-quality datasets, such as LIMO or s1k, can outperform larger ones: consistent supervision leads to more effective optimization.

\section{Conclusion}
We constructed a high-quality reverse reasoning dataset r1k and demonstrated its effectiveness in improving reasoning ability. We further investigate the effects of mixed data during multi-stage fine-tuning, underscoring the need for improved alignment strategies to support robust reasoning.

\section{Limitations}
This work explores the integration of reverse reasoning data in multi-stage fine-tuning, but several limitations remain. Our reverse dataset \(\mathcal{D}_{r1k}\) is constructed by automated prompting without human validation, which may introduce subtle errors or inconsistencies in reasoning quality. Additionally, the DPO formulation assumes a strict directional preference between forward and reverse outputs, potentially oversimplifying cases where both reasoning directions offer complementary insights. Furthermore, while this study adopts a standard SFT + DPO pipeline, alternative alignment strategies may offer more robust solutions to conflicting supervision in mixed data settings.

\section{Ethics Statement}  
This work trains and evaluates models on publicly available datasets (s1k, GPQA, AIME, and Math~500), each used in accordance with its respective license. We release our \(\mathcal{D}_{r1k}\) dataset and code strictly for research purposes. In this study, no personal or sensitive information is collected or used.

\section{Acknowledgements}   
We would like to sincerely thank the anonymous reviewers and area chairs for their valuable comments and constructive feedback, which have substantially improved this work. This research was supported in part by the Guangdong Provincial Key Lab of Integrated Communication, Sensing and Computation for Ubiquitous Internet of Things (No.\ 2023B1212010007, SL2023A03J00934), and by the Guangzhou Municipal Science and Technology Project (No.\ 2023A03J0003, 2023A03J0013, and 2024A03J0621).

\bibliography{custom}

\appendix
\section{Appendix}
\subsection{Reverse Data Construction}
\newtcolorbox{promptbox}{
  colback=gray!5,
  colframe=gray!50,
  boxrule=0.5pt,
  arc=2pt,
  left=5pt,
  right=5pt,
  top=5pt,
  bottom=5pt,
  fontupper=\ttfamily\footnotesize,
  enhanced,
  breakable 
}

We design two prompt templates to support our bidirectional reasoning framework. These templates serve complementary purposes in data construction, enabling the model to learn both forward and reverse reasoning patterns. The first template is designed to generate reverse question-answer (QA) pairs. Given a question and its corresponding answer, the model is instructed to invert their roles: reformulating the answer into a thought-provoking question and the original question into a reflective statement that serves as the new answer.

\begin{promptbox}
"role\_definition": \\
"You are an AI model tasked with generating a reflective thinking exercise. \\
Given the following question and answer:" \\

- \textbf{Question}: \{question\} \\
- \textbf{Answer}: \{answer\} \\

"instructions": \\
"Your task is to reverse the roles of the question and answer. \\
Transform the answer into a question that is thought-provoking and encourages deeper reflection. \\
Similarly, convert the original question into a statement that serves as an insightful answer. \\
Ensure that the new question remains reasonable and stimulates further inquiry, \\
while the new answer is right to the question." \\

"expected\_output": \\
- \textbf{New Question}: \\
- \textbf{New Answer}:
\end{promptbox}

The second template focuses on generating step-by-step solutions with CoT prompting, followed by a clearly demarcated final answer. This template is specifically designed to extract DeepSeek R1’s detailed reasoning chains and final answers, serving as the foundation for producing high-quality reasoning examples.

\begin{promptbox}
"role\_definition": \\
"You are an AI model that is designed to generate solutions to a given question. \\
All numerical answers must be explicitly marked with \textbackslash boxed\{\}." \\[0.5em]

- \textbf{Question}: \{question\} \\[0.5em]

"instructions": \\
"Ensure your answer is absolutely correct and standard." \\[0.5em]

"expected\_output": \\
Presents the complete and concise answer. \\
If the answer contains only one numerical value, it must be marked in the form of \textbackslash boxed\{\}.
\end{promptbox}

\begin{figure*}[t!] 
    \centering
    \includegraphics[width=1\linewidth]{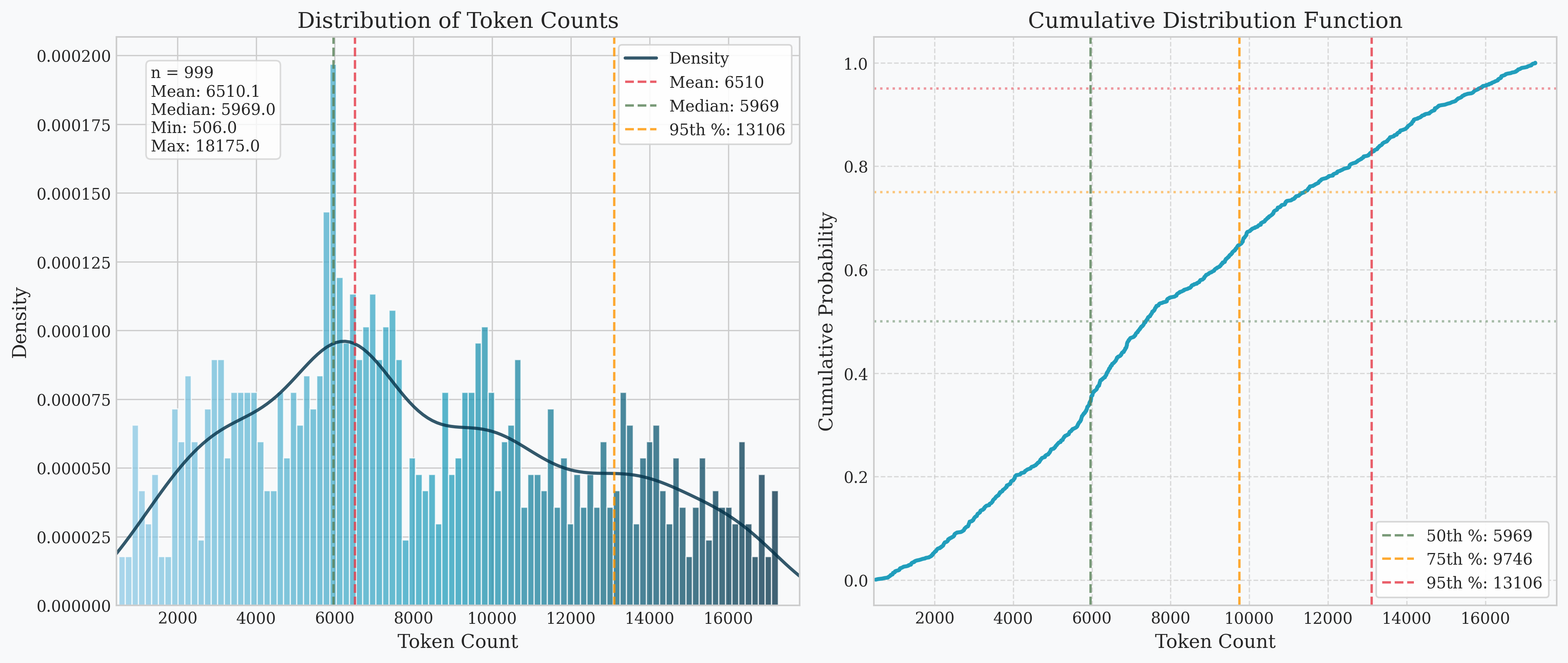}
    \caption{Distribution of r1k Token Counts}
    \label{tokens}
\end{figure*}
\begin{figure*}[t!] 
    \centering
    \includegraphics[width=1\linewidth]{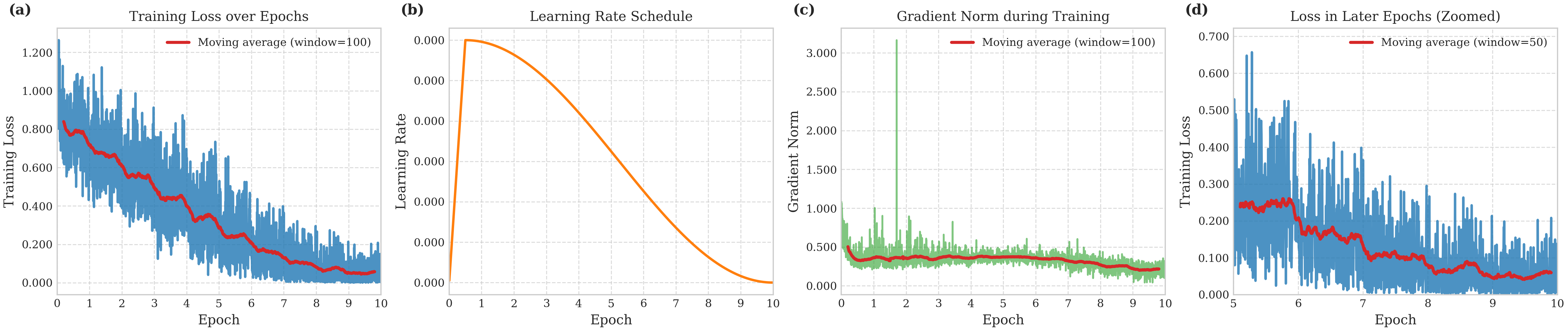}
    \caption{SFT Mixed data Training Details}
    \label{3}
\end{figure*}


\begin{figure*}[t!]
    \centering
    \includegraphics[width=1\linewidth]{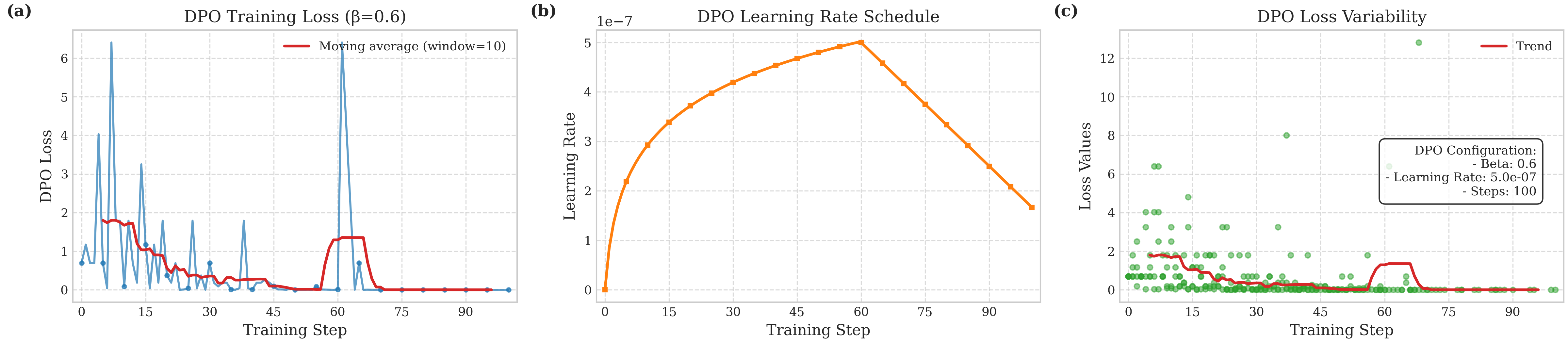}
    \caption{DPO-Training Details (First 100 steps)}
    \label{fig:5}
\end{figure*}

The r1k dataset shares the same domains as the s1k dataset~\cite{muennighoff2025s1}, and its token length distribution is shown in Figure~\ref{tokens}. We select the 50 pieces of data with the least text tokens in s1k and the corresponding r1k data as our probe dataset. All external datasets used in our work are publicly available and licensed for research use. Specifically, AIME and OpenAI's Math 500 datasets are distributed under the Apache License 2.0, while GPQA is released under the Creative Commons Attribution 4.0 International (CC BY 4.0) license.

\subsection{Training Details}

The training curves in Figures~\ref{3}–\ref{fig:5} illustrate learning dynamics under different supervision settings by tracking key optimization parameters of loss, learning rate, and gradient norm. Figure~\ref{3} depicts the progression of SFT on mixed data, while Figure~\ref{4} provides a probe-based analysis across forward, reverse, and mixed reasoning styles. Figure~\ref{fig:5} captures the early-stage DPO optimization, emphasizing how preference alignment evolves within the first 100 training steps. 

\subsection{Additional Experimental Results}

\subsubsection{DPO with Separate SFT Stage}
In addition to the SFT and DPO pipeline, we also evaluate applying DPO directly 
on the mixed dataset $\mathcal{D}$ (2k examples from s1k and r1k), without a prior SFT stage. 
Results in Table~\ref{tab:appendix_direct_dpo} show that direct DPO yields lower performance 
than SFT followed by DPO, confirming that SFT provides a stable initialization.

\begin{table}[h]
\centering
\small
\setlength{\abovecaptionskip}{4pt}
\setlength{\belowcaptionskip}{0pt}
\caption{Direct DPO (without SFT) on mixed datasets $\mathcal{D}$ (2k data from s1k and r1k).}
\setlength{\tabcolsep}{3.5pt}
\begin{tabular}{ccccc}
\toprule
\textbf{Model} & \textbf{AIME} & \textbf{GPQA} & \textbf{Math} & \textbf{Average} \\
\midrule
7B  & 10.0\% & 34.6\% & 59.0\% & 34.53\% \\
14B & 20.0\% & 40.2\% & 72.0\% & 44.07\% \\
\bottomrule
\end{tabular}
\label{tab:appendix_direct_dpo}
\end{table}



\subsection{Experiment Settings}

We adopt a two-stage training pipeline consisting of supervised SFT followed by DPO. This section details our model configurations, optimization strategies, and resource settings.

\subsubsection{Supervised Fine-Tuning (SFT)}

We use the \texttt{Qwen2.5-7B-Instruct} model and train on 4 A800 GPUs, with a maximum sequence length of 20{,}000 tokens. Some of the excess data are replaced with Gemini data for compatibility. For the \texttt{Qwen2.5-14B-Instruct} model, we performed training on 8 A800 GPUs under the same sequence-length setting. Training 7B-based models with 1k data takes around 3 hours, while 14B-based models requires 6-7 hours.

We adopt a cosine learning rate schedule with restarts and set the initial learning rate to $3\times10^{-4}$. Other hyperparameters include:
\begin{itemize}
    \item Weight decay: $1\times10^{-6}$
    \item Adam optimizer: $\beta_1 = 0.9$, $\beta_2 = 0.95$
    \item Gradient clipping: 1.0
    \item Warmup ratio: 0.05
\end{itemize}

\paragraph{Training Configuration.}
\begin{itemize}
    \item Number of epochs: 10
    \item Per-device batch size: 1
    \item Gradient accumulation steps: 1
    \item Mixed precision: BF16
\end{itemize}

\paragraph{Parameter-Efficient Fine-Tuning.}

We apply LoRA with the following configuration:
\begin{itemize}
    \item Rank $r = 256$, scaling factor $\alpha = 512$
    \item Target modules: \texttt{"q\_proj, k\_proj, v\_proj, o\_proj, gate\_proj, down\_proj, up\_proj, lm\_head"}
    \item Optimizer: 8-bit AdamW
\end{itemize}

\subsubsection{Direct Preference Optimization (DPO)}

We initialize from the SFT checkpoint and train on 4 GPUs. The training runs for a maximum of 200 steps in around 3 hours. We experiment with DPO using the value of $\beta$ ranging from 0.1 to 0.8, and observe that $\beta{=}0.6$ leads to the most significant performance improvement on mixed-direction data. However, under this setting, the change in average log-probability (ALP) is relatively subtle. To better highlight learning dynamics in our probe-based analysis, we use a smaller $\beta{=}0.2$, which yields more distinguishable ALP shifts between $y^+$ and $y^-$.

\paragraph{Optimization Settings.}
\begin{itemize}
    \item Learning rate: $5\times10^{-7}$ with cosine decay
    \item KL penalty coefficient $\beta = 0.6$ for DPO training, $\beta = 0.2$ for probe dataset training and testing.
\end{itemize}

\paragraph{Training Configuration.}
\begin{itemize}
    \item Per-device batch size: 1
    \item Gradient accumulation: 4 steps
    \item Mixed precision: BF16
    \item Maximum prompt length: 600 tokens
    \item Maximum sequence length: 20{,}000 tokens
\end{itemize}

\paragraph{Efficiency Enhancements.}
We enable the following optimizations to reduce memory and computation overhead:
\begin{itemize}
    \item Flash Attention
    \item Gradient checkpointing
    \item DeepSpeed ZeRO-3 with parameter sharding
\end{itemize}

\paragraph{Distributed Optimization.}
We apply DeepSpeed ZeRO-3 for memory efficiency and optimize multi-GPU communication through NCCL tuning.

\subsection{Example of Forward–Reverse Data Construction}
\label{sec:example}

As an illustrative case, we present a representative forward–reverse pair from $\mathcal{D}_{s1k}$ and $\mathcal{D}_{r1k}$.
The forward instance is taken directly from $\mathcal{D}_{s1k}$, whereas the reverse instance is obtained by inverting the question–answer direction and generating the corresponding chain-of-thought with DeepSeek-R1.

\subsubsection{Forward Example (from \texorpdfstring{$\mathcal{D}_{s1k}$}{Ds1k}).}

\textit{Question:} Find any quadruple of positive integers $(a, b, c, d)$ satisfying $a^{3}+b^{4}+c^{5}=d^{11}$ and $a b c<10^{5}$.  

\textit{Answer:} It's easy to guess that there are solutions such that $a, b, c, d$ are in the form of $n^{x}$, where $n$ is a rather small number. After a few attempts, we can see that we obtain simple equations when $n=2$ or $n=3$ : for $n=2$, the equation becomes in the form of $2^{t}+2^{t}+2^{t+1}=2^{t+2}$ for some non-negative integer $t$; for $n=3$, the equation becomes in the form of $3^{t}+3^{t}+3^{t}=3^{t+1}$ for some non-negative integer $t$. In the first case, we hope that $t$ is a multiple of two of $3,4,5$, that $t+1$ is a multiple of the last one, and that $t+2$ is a multiple of 11 . Therefore, $t \equiv 15,20,24(\bmod 60)$ and $t \equiv 9(\bmod 11)$. It's easy to check that the only solution that satisfies the given inequality is the solution with $t=20$, and $(a, b, c, d)=(128,32,16,4)$. In the case where $n=3$, we must have that $t$ is a multiple of 60 , which obviously doesn't satisfy the inequality restriction. Remark: By programming, we find that the only two solutions are $(a, b, c, d)=(128,32,16,4)$ and $(a, b, c, d)=(160,16,8,4)$, with the the former being the intended solution.  

\subsubsection{Reverse Example (from \texorpdfstring{$\mathcal{D}_{r1k}$}{Dr1k}).}

\textit{Question:} The quadruples (128, 32, 16, 4) and (160, 16, 8, 4) satisfy a specific equation involving exponents of the form \(a^k + b^m + c^n = d^p\), along with the condition \(abc < 10^5\). What is the exact equation (including the exponents \(k, m, n, p\)) and the constraint? 

\textit{Answer:} The equation satisfied by both quadruples is \(a^3 + b^4 + c^5 = 4^{11}\) with the constraint \(abc < 10^5\). 
For the first quadruple \((128, 32, 16, 4)\): 
- \(128^3 + 32^4 + 16^5 = 2^{21} + 2^{20} + 2^{20} = 2^{22} = 4^{11}\). 
- \(abc = 128 \times 32 \times 16 = 65,\!536 < 10^5\).
For the second quadruple \((160, 16, 8, 4)\): 
- \(160^3 + 16^4 + 8^5 = (16 \times 10)^3 + 2^{16} + 2^{15} = 4,\!096,\!000 + 65,\!536 + 32,\!768 = 4,\!194,\!304 = 4^{11}\). 
- \(abc = 160 \times 16 \times 8 = 20,\!480 < 10^5\).
Thus, the exact equation is \(\boxed{a^3 + b^4 + c^5 = 4^{11}}\) with the constraint \(abc < 10^5\).  

\textit{Chain-of-thought (concise overview):}

Observe $128=2^7,\,32=2^5,\,16=2^4,\,4=2^2$.  
Enforce $5m=4n$ so the latter two terms match in exponent; taking $m=4,\,n=5$ gives both as $2^{20}$.  
Choose $k=3$ so $2^{7k}=2^{21}$, and the sum becomes $2^{21}+2\cdot 2^{20}=2^{22}=4^{11}$, hence $p=11$.  
The same exponents verify the second quadruple.

\vspace{1ex}

\end{document}